\documentclass[10pt,twocolumn,letterpaper]{article}

\usepackage{cvpr}
\usepackage{times}
\usepackage{epsfig}
\usepackage{graphicx}
\usepackage{amsmath}
\usepackage{amssymb}

\usepackage{multirow}
\usepackage{multicol}
\usepackage{longtable}
\usepackage{float}
\usepackage{diagbox}
\usepackage{stmaryrd}
\usepackage{caption} 

\usepackage[hidelinks]{hyperref}
\usepackage{xcolor}
\definecolor{hyperlink}{rgb}{10,0,0}

\cvprfinalcopy 

\def\cvprPaperID{3194} 

\begin{document}

\title{DPGN: Distribution Propagation Graph  Network for Few-shot Learning}


\author{Ling Yang\textsuperscript{1}\thanks{Contributed equally.}\ \ \ \  Liangliang Li\textsuperscript{2}\footnotemark[1]\ \    \thanks{Corresponding author.}  \ \ \  Zilun Zhang\textsuperscript{2} \ \ \ Xinyu Zhou\textsuperscript{2} \ \ Erjin Zhou\textsuperscript{2}\ \ \ Yu Liu\textsuperscript{2}\\
Northwestern Polytechnical University\textsuperscript{1}\ \ \ \ Megvii Technology\textsuperscript{2}\\
{\tt\small yangling0818@163.com,\ liliangliang@megvii.com} \\
{\tt\small \ zilun.zhang@mail.utoronto.ca,  \{zxy, zej, liuyu\}@megvii.com}}


\maketitle
\begin{abstract}
    
Most graph-network-based meta-learning approaches model instance-level relation of examples.   
We extend this idea further to explicitly model the distribution-level relation of one example to all other examples in a 1-vs-N manner. 
We propose a novel approach named distribution propagation graph network (DPGN) for few-shot learning. It conveys both the distribution-level relations and  instance-level relations in each few-shot learning task.
To combine the distribution-level  relations and instance-level relations for all examples, we construct a dual complete graph network which consists of a point graph and a distribution graph with each node standing for an example. Equipped with dual graph architecture, DPGN propagates label information from labeled examples to unlabeled examples within several update generations.
In extensive experiments on few-shot learning benchmarks, DPGN outperforms state-of-the-art results by a large margin in 5\% $\sim$ 12\% under supervised settings and 7\% $\sim$ 13\%  under semi-supervised settings. Code is available at \href{https://github.com/megvii-research/DPGN}{\textcolor{hyperlink}{https://github.com/megvii-research/DPGN}}
\end{abstract}

\section{Introduction}
The success of deep learning is rooted in a large amount of labeled data~\cite{krizhevsky2012imagenet,sun2017revisiting}, while humans generalize well after having seen few examples. The contradiction between these two facts brings great attention to the research of few-shot learning~\cite{fei2006one,lake2011one}. Few-shot learning task aims at predicting unlabeled data (query set) given a few labeled data (support set).

\begin{figure}[h]

\centering

    \includegraphics[width=8cm]{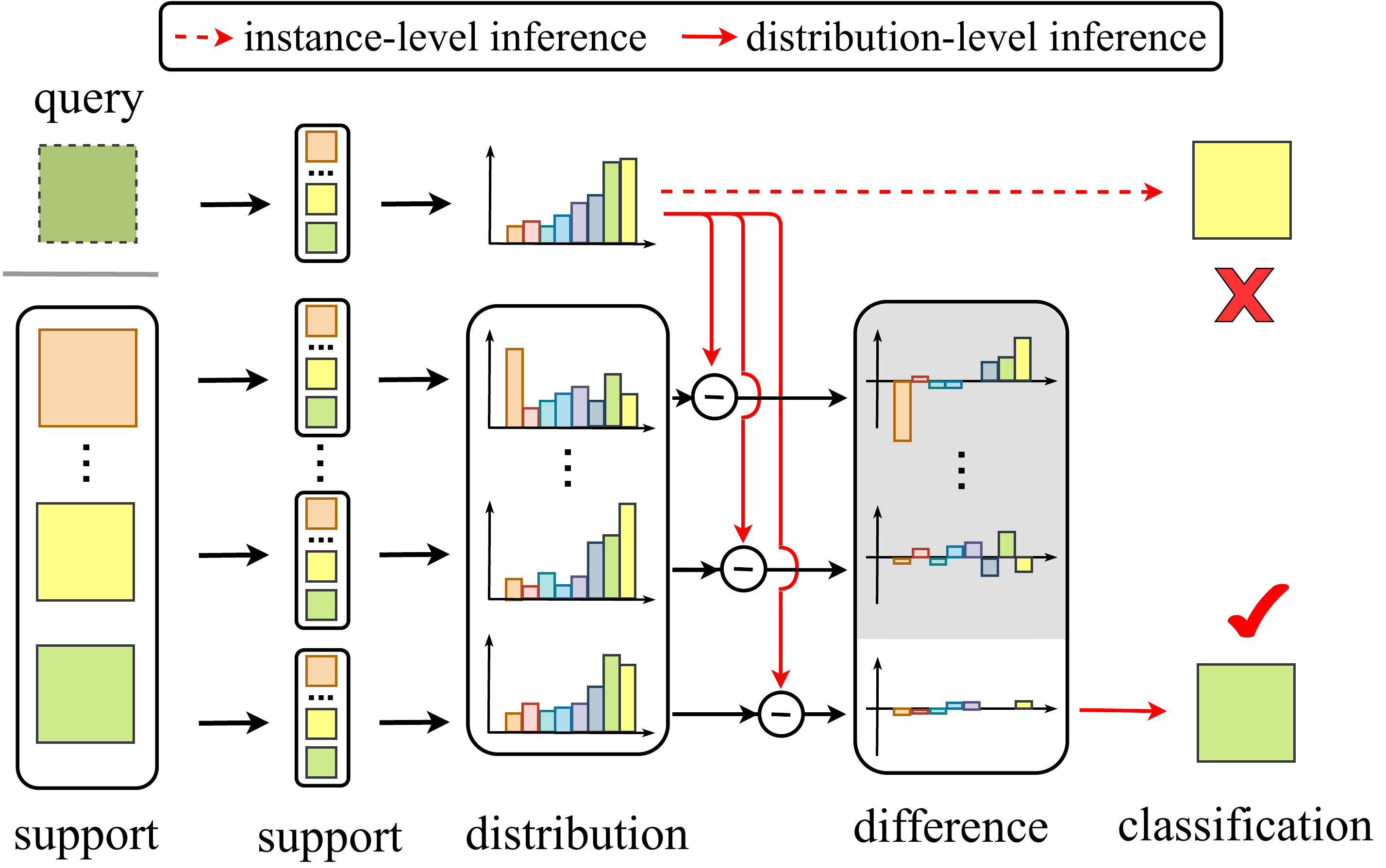}


\caption{Our proposed DPGN adopts contrastive comparisons between each sample with support samples to produce distribution representation.  Then it incorporates distribution-level comparisons with instance-level comparisons when classifying the query sample.}
\label{fig:f0}
\end{figure}

Fine-tuning~\cite{chen19closerfewshot} is the defacto method in obtaining a predictive model from a small training dataset in practice nowadays. However, it suffers from overfitting issues~\cite{gidaris2018dynamic}.
Meta-learning~\cite{finn2017model} methods introduces the concept of \textit{episode} to address the few-shot problem explicitly. An episode is one round of model training, where in each episode, only few examples (e.g., 1 or 5) are randomly sampled from each class in training data. Meta-learning methods adopt a trainer (also called meta-learner) which takes the few-shot training data and outputs a classifier. This process is called \textit{episodic training}~\cite{vinyals2016matching}.
Under the framework of meta-learning, a diverse hypothesis was made to build an efficient meta-learner.

A rising trend in recent researches was to process the training data with 
Graph Networks \cite{ battaglia2018relational}, which is a powerful model that generalizes many data structures (list, trees) while introduces a combinatorial prior over data. Few-Shot GNN \cite{garcia2017few} is proposed to build a complete graph network where each node feature is concatenated with the corresponding class label, then node features are updated via the attention mechanism of graph network to propagate the label information. 
To further exploit intra-cluster similarity and inter-cluster dissimilarity in the graph-based network, EGNN \cite{kim2019edge} demonstrates an edge-labeling graph neural network under the \textit{episodic training} framework.
It is noted that previous GNN studies in few-shot learning mainly focused on pair-wise relations like node labeling or edge labeling, and ignored a large number of substantial distribution relations.
Additionally, other meta-learning approaches claim to make use of the benefits of global relations by \textit{episodic training}, but in an implicitly way.

As illustrated in Figure \ref{fig:f0}, firstly,  we extract the \textit{instance feature} of support and query samples. Then, we obtain the \textit{distribution feature} for each sample by calculating the instance-level similarity over all support samples.
To leverage both instance-level and  distribution-level representation of each example and process the representations at different levels independently, we propose a dual-graph architecture: a point graph (PG) and a distribution graph (DG). 
Specifically, a PG generates a DG by gathering 1-vs-n relation on every example, while the DG refines the PG by delivering distribution relations between each pair of examples. Such cyclic transformation adequately fuses instance-level and distribution-level relations and multiple generations (rounds) of this Gather-Compare process concludes our approach.
Furthermore, it is easy to extend DPGN to semi-supervised few-shot learning task where support set containing both labeled and unlabeled samples for each class. DPGN builds a bridge connection between labeled and unlabeled samples in the form of similarity distribution, which leads to a better propagation for label information in semi-supervised few-shot classification.





Our main contributions are summarized as follows:
\begin{itemize}
\item To the best of our knowledge, DPGN is the first to \textbf{explicitly incorporate distribution propagation} in graph network for few-shot learning. The further ablation studies have demonstrated the effectiveness of distribution relations. 
\item We devise the \textbf{dual complete graph network} that combines instance-level and distribution-level relations. The cyclic update policy in this framework contributes to enhancing instance features with distribution information.
\item Extensive experiments are conducted on four popular benchmark datasets for few-shot learning. By comparing with all state-of-the-art methods, the DPGN achieves a significant improvement of \textbf{5\%$\sim$12\%} on average in few-shot classification accuracy. 
In semi-supervised tasks, our algorithm outperforms existing graph-based few-shot learning methods by \textbf{7\%$\sim$13 \%}.
\end{itemize}

\section{Related Work}

\subsection{Graph Neural Network} Graph neural networks were first designed for tasks on processing graph-structured data \cite{scarselli2008graph,vinyals2016matching}. 
Graph neural networks mainly refine the node representations by aggregating and transforming neighboring nodes recursively. 
Recent approaches \cite{garcia2017few,liu2018learning,kim2019edge} are proposed to exploit GNN in the field of few-shot learning task. 
TPN \cite{liu2018learning} brings the transductive setting into graph-based few-shot learning, which performs a Laplacian matrix to propagate labels from support set to query set in the graph.  
It also considers the similarity between support and query samples through the process of pairwise node features affinities to propagate labels.
EGNN \cite{kim2019edge} uses the similarity/dissimilarity between samples and dynamically update both node and edge features for complicated interactions.

\subsection{Metric Learning}
Another category of few-shot learning approaches focus on optimizing feature embeddings of input data using metric learning methods.
Matching Networks \cite{vinyals2016matching} produces a weighted nearest neighbor classifier through computing embedding distance between support and query set.
Prototypical Networks \cite{snell2017prototypical} firstly build a prototype representation of each class in the embedding space.
As an extension of Prototypical Networks, IMF \cite{allen2019infinite} constructs infinite mixture prototypes by self-adaptation. 
RelationNet \cite{sung2018learning} adopts a distance metric network to learn pointwise relations in support and query samples. 

\begin{figure*}[h]
\centering
    \includegraphics[width=17.5cm]{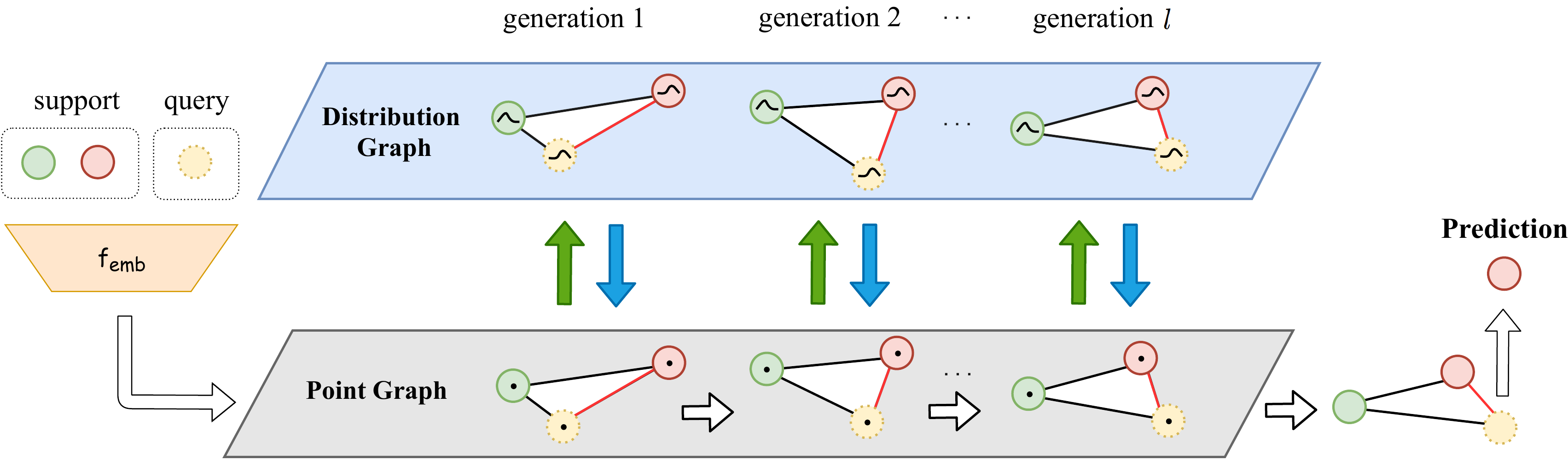}


\caption{The overall framework of DPGN. In this illustration, we take a 2way-1shot task as an example. The support and query embeddings obtained from feature extractor are delivered to the dual complete graph (a point graph and a distribution graph) for transductive propagation generation after generation. The \textbf{green arrow} represents a edge-to-node transformation ($\operatorname{P2D}$, described in Section \ref{sec:1}) which aggregates instance similarities to construct distribution representations and the \textbf{blue arrow} represents another edge-to-node transformation ($\operatorname{D2P}$, described in  Section \ref{sec:2}) which aggregates distribution similarities with instance features. DPGN makes the prediction for the query sample at the end of generation $l$.} 
\label{fig:f1}
\end{figure*}

\subsection{Distribution Learning}
Distribution Learning theory was first introduced in~\cite{kearns1994learnability} to find an efficient algorithm that determines the distribution from which the samples are drawn. Different methods~\cite{kalai2010efficiently,dasgupta1999learning,daskalakis2015learning} are proposed to efficiently estimate the target distributions. DLDL~\cite{gao2017deep} is one of the researches that has assigned the discrete distribution instead of one-hot label for each instance in classification and regression tasks. CPNN~\cite{yin2012facial} takes both features and labels as the inputs and produces the label distribution with only one hidden layer in its framework. LDLFs~\cite{shen2017label} devises a distribution learning method based on the decision tree algorithm.

\subsection{Meta Learning}
Some few-shot approaches adopt a meta-learning framework that learns meta-level
knowledge across batches of tasks. MAML~\cite{finn2017model} are gradient-based approaches that design the meta-learner as an optimizer that could learn to update the model parameters (e.g., all layers of a deep network) within few optimization steps given novel examples. Reptile~\cite{nichol2018first} simplifies the computation of meta-loss by incorporating an L2 loss which updates the meta-model parameters towards the instance-specific adapted models. SNAIL~\cite{mishra2017simple} learn a parameterized predictor to estimate the parameters in models. MetaOptNet~\cite{lee2019meta} advocates the use of linear classifier instead of nearest-neighbor methods which can be optimized as convex learning problems. LEO~\cite{rusu2018meta} utilizes an encoder-decoder architecture to mine the latent generative representations and predicts high-dimensional parameters in extreme low-data regimes.

\section{Method}

In this section, we first provide the background of few-shot learning task, then introduce the proposed algorithm in detail.

\subsection{Problem Definition}

The goal of few-shot learning tasks is to train a model that can perform well in the case where only few samples are given. 

Each few-shot task has a \emph{support set} $\mathcal{S}$ and a \emph{query set} $\mathcal{Q}$. Given training data $\mathbb{D}^{train}$, 
the support set $\mathcal{S} \subset \mathbb{D}^{train}$ contains $N$ classes with $K$ samples for each class (i.e., the $N$-way $K$-shot setting), it can be denoted as $\mathcal{S} = \{(x_{_1},y_{_1}),(x_{_2},y_{_2}),\dots,(x_{{_N} {_\times} {_K}},y_{{_N} {_\times} {_K}})\}$.
The \emph{query} set $\mathcal{Q} \subset \mathbb{D}^{train}$ has $\bar{T}$ samples and can be denoted as $\mathcal{Q} = \{(x_{_{N \times K+1}},y_{_{N \times K+1}}),\dots,(x_{_{N \times K+\bar{T}}},y_{_{N \times K+\bar{T}}})\}$.
Specifically, in the training stage, data labels are provided for both support set $\mathcal{S}$ and query set $\mathcal{Q}$.
Given testing data $\mathbb{D}^{test}$, our goal is to train a classifier that can map the query sample from $\mathcal{Q} \in \mathbb{D}^{test}$   to the corresponding label accurately with few support samples from $\mathcal{S} \in \mathbb{D}^{test}$. Labels of support sets and query sets are mutually exclusive.


\subsection{Distribution Propagation Graph Networks}

In this section, we will explain the DPGN that we proposed for few-shot learning in detail. As shown in Figure \ref{fig:f1}.
The DPGN consists of $l$ generations and each generation consists of a point graph $G^p_l = (V^p_l, E^p_l)$ and a distribution graph $G^d_l = (V^d_l, E^d_l)$. Firstly, the feature embeddings of all samples are extracted by a convolutional backbone, these embeddings are used to compute the instance similarities $E^p_l$. Secondly, the instance relations $E^p_l$ are delivered to construct the distribution graph $G^d_l$. The node features $V^d_l$ are initialized by aggregating $E^p_l$ following the position order in $G^p_l$ and the edge features $E^d_l$ stand for the distribution similarities between the node features $V^d_l$. Finally,  the obtained $E^d_l$ is delivered to $G^p_l$ for constructing more discriminative representations of nodes $V^p_l$ and we repeat the above procedure generation by generation. A brief introduction of generation update for the DPGN can be expressed as $E^p_l\xrightarrow{}V^{d}_{l}\xrightarrow{}E^d_l\xrightarrow{}V^{p}_{l}\xrightarrow{}E^p_{l+1}$, where $l$ denotes the $l$-th generation.

For further explanation, we formulate $V^{p}_{l}$, $E^p_l$, $V^{d}_{l}$ and $E^d_l$ as follows:\ $V^{p}_{l} = \{v^{p}_{l,i}\}$,  
$E^p_l= \{e^p_{l,ij}\}$, 
${V}^d_l= \{v^d_{l,i}\}$, 
$E^d_l = \{e^d_{l,ij}\}$ where $i, j = 1,\cdots T$.
$T = N\times K + \bar{T}$ denotes the total number of examples in a training episode. $v^p_{0, i}$ is first initialized by the output of the feature extractor $f_{emb}$. For each sample $x_i$:
\begin{align}
 v^{p}_{0,i} = f_{emb}(x_i) \,,
\end{align}
where $v^p_{0, i} \in \mathbb{R}^m$ and $m$ denotes the dimension of the feature embedding.

\begin{figure}[h]

\centering

    \includegraphics[width=8.5cm]{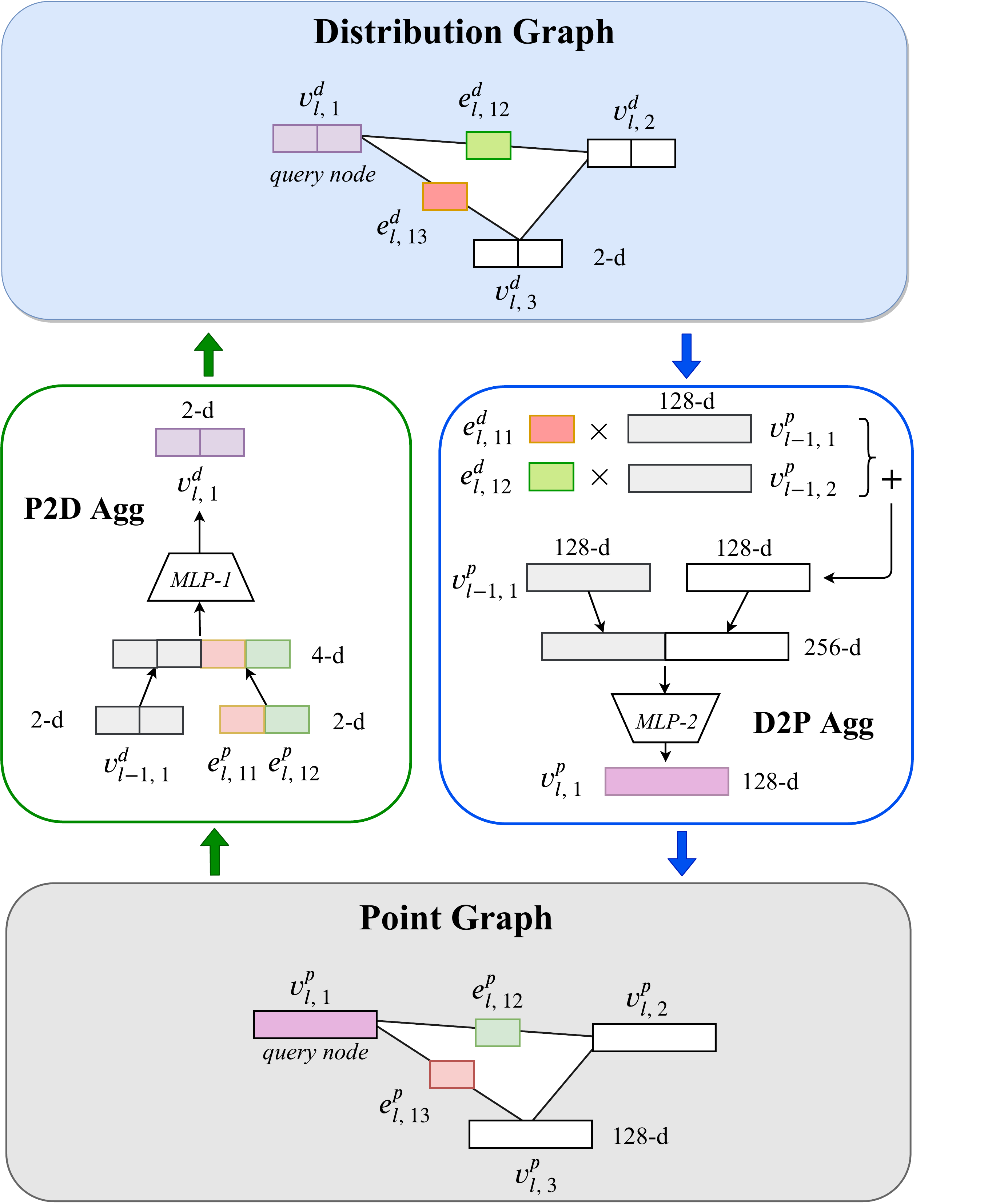}


\caption{Details about \textbf{P2D aggregation} and \textbf{D2P aggregation} in DPGN. A 2way-1shot task is presented as an example. MLP-1 is the FC-ReLU blocks mentioned in P2D Aggregation and MLP-2 is the Conv-BN-ReLU blocks mentioned in D2P Aggregation. The green arrow denotes the P2D aggregation while the blue arrow denotes the D2P aggregation. Both aggregation processes integrate the node or edge features of their previous generation.}
\label{fig:f8}
\end{figure}

\subsubsection{Point-to-Distribution Aggregation}
\paragraph{Point Similarity} Each edge in the point graph stands for the instance (point) similarity and the edge  ${e^p_{0,ij}}$ of the first generation is initialized as follows:
\begin{align} 
e^p_{0,ij} = {f_{e^p_{0}} (({v^p_{0,i}}-{v^p_{0,j}})^2)}\,,
\label{eq:2}
\end{align}
where $e^p_{0,ij} \in \mathbb{R}$. $f_{e^p_{0}}:\mathbb{R}^m\xrightarrow{}\mathbb{R}$ is the encoding network that transforms the instance similarity to a certain scale. $f_{e^p_{0}}$ contains two Conv-BN-ReLU \cite{glorot2011deep, ioffe2015batch} blocks with the parameter set ${\theta}_{e^p_{0}}$ and a sigmoid layer.

For generation $l>0$, given $e^p_{l-1, ij}$, $v^p_{l-1, i}$ and $v^p_{l-1, j}$, $e^p_{l, ij}$ can be updated as follows:
\begin{align}
    {e^p_{l,ij}} = {f_{e^p_l} (({v^p_{l-1,i}}-{v^p_{l-1,j}})^2) \cdot {e^p_{l-1,ij}}}
    \label{eq:3}\,.
\end{align}
In order to use edge information with a holistic view of the graph $G^p_l$, a normalization operation is conducted on the ${e^p_{l,ij}}$.

\paragraph{P2D Aggregation} \label{sec:1} After edge features $E^p_l$ in point graph $G^p_l$ are produced or updated, the distribution graph $G^d_l= ({V}^d_l, E^d_l$) is the next to be constructed. As shown in Figure \ref{fig:f8},
$G^d_l$ aims at integrating instance relations from the point graph $G^p_l$ and process the \textit{distribution-level} relations. 
Each distribution feature $v^d_{l,i}$ in $G^d_l$ is a $N K$ dimension feature vector  where the value in $j$-th entry represents the relation between sample $x_i$ and sample $x_j$ and $N K$ stands for the total number of support samples in a task. 
For first initialization: 
\begin{align} v^d_{0,i} =
\begin{cases}
\bigparallel^{N K}_{j=1} \delta(y_i, y_j) & \text{if}\ x_i \ \text{is labeled}, \\
[\frac{1}{N K},\cdots,\frac{1}{N K} ] &  \text{otherwise} , 
\end{cases}
\end{align}
where $v^d_{0, i} \in \mathbb{R}^{N  K}$ and $\bigparallel$ is the concatenation operator. $\delta(\cdot)$ is the Kronecker delta  function which outputs one when $y_i = y_j$ and zero otherwise ($y_i$ and $y_j$ are labels).
 
For generations $l>0$, the distribution node $ {v^{d}_{l,i}}$ can be updated as follows: 
\begin{align}
    v^d_{l,i} =  \operatorname{P2D} (\bigparallel_{j=1}^{N K} e^p_{l,ij}\ , \ v^d_{l-1,i} )\,,
    \label{eq:5}
\end{align}
where $\operatorname{P2D}$ $:(\mathbb{R}^{N K},\ \mathbb{R}^{N K})\xrightarrow{}\mathbb{R}^{N K}$ is the aggregation network for distribution graph. $\operatorname{P2D}$ applies a  concatenation operation between two features. Then, P2D performs a transformation $:\mathbb{R}^{2 N K}\xrightarrow{}\mathbb{R}^{N K}$ on the concatenated features which is composed of a fully-connected layer and ReLU \cite{glorot2011deep}, with the parameter set ${\theta}_{v^d_{l}}$.

\subsubsection{Distribution-to-Point Aggregation}

\paragraph{Distribution Similarity} Each edge in distribution graph stands for the similarity between distribution features of different samples. For generation $l=0$, the distribution similarity ${e^d_{0,ij}}$ is initialized as follows:
\begin{align}
    {e^d_{0,ij}} = {f_{e^d_{0}} (({v^d_{0,i}}-{v^d_{0,j}})^2)}\,,
    \label{eq:6}
\end{align}
where $e^d_{0,ij} \in \mathbb{R}$. The encoding network $f_{e^d_{0}}:\mathbb{R}^{N K}\xrightarrow{}\mathbb{R}$ transforms the distribution similarity using two Conv-BN-ReLU blocks with the parameter set ${\theta}_{e^d_{0}}$ and a sigmoid layer in the end.
For generation $l>0$, the update rule for $e^d_{l, ij}$ in $G^{d}_{l}$ is formulated as follows:
\begin{align}
    {e^d_{l,ij}} = {f_{e^d_{l}}( ({v^d_{l,i}}-{v^d_{l,j}})^2) \cdot {e^d_{l-1,ij}}}\,.
    \label{eq:7}
\end{align}
Also, we apply a normalization to $e^d_{l, ij}$.

\paragraph{D2P Aggregation} \label{sec:2} As illustrated in Figure \ref{fig:f8}, the encoded distribution information in $G^d_l$ flows back into the point graph $G^p_l$ at the end of each generation. Then node features ${v}^p_{l, i}$ in $G^p_l$ captures the distribution relations through aggregating all the node features in $G^p_l$ with edge features ${e}^d_{l, i}$ as follows:  
\begin{align}
     v^p_{l,i} =  \operatorname{D2P}(\sum_{j=1}^{T} (e^d_{l,ij}\cdot v^p_{l-1,j}), v^p_{l-1,i})\,,
     \label{eq:8}
\end{align}
where $v^p_{l, i} \in \mathbb{R}^{m}$ and $\operatorname{D2P} : (\mathbb{R}^{m}, \mathbb{R}^{m})\xrightarrow{}\mathbb{R}^{m}$ is the aggregation network for point graph in $G^p_l$ with the parameter set $\theta_{v^p_{l}}$. 
$\operatorname{D2P}$ concatenates the feature which is computed by $\sum_{j=1}^T (e^d_{l,ij}\ \cdot \ v^p_{l-1, j})$ with the node features ${v}^p_{l-1, i}$ in previous generation and update the concatenated feature with two Conv-BN-ReLU blocks. After this process, the node features can integrate the distribution-level information into the instance-level feature and prepares for computing instance similarities in the next generation.

\subsection{Objective}

The class prediction of each node  can be computed by feeding the corresponding edges in the final generation $l$ of DPGN into softmax function:
\begin{align}
    P( \hat{y_i} | x_i)  = \operatorname{Softmax}  (\sum_{j=1}^{N K} {e^{p}_{l,ij}} \cdot \operatorname{\textit{one-hot}}(y_j))\,,
\end{align}
where $P( \hat{y_i} | x_i)$ is the probability distribution over classes given sample $x_i$, and $y_j$ is the label of $j$th sample in the support set. $e^p_{l,ij}$ stands for the edge feature in the point graph at the final generation.

\paragraph{Point Loss} 
It is noted that we make classification predictions in the point graph for each sample. Therefore, \textit{the point loss} at generation $l$ is defined as follows:

\begin{align}
    \mathcal{L}_l^p = \mathcal{L}_{CE}(P( \hat{y_i} | x_i) , {y_i})\,,
\end{align}
where $\mathcal{L}_{CE}$ is the cross-entropy loss function, $T$ stands for the number of samples in each task $(S,Q) \in D^{train}$.  $P( \hat{y_i} | x_i)$ and ${y_i}$ are model probability predictions of sample $x_i$ and the ground-truth label respectively.

\paragraph{Distribution Loss} To facilitate the training process and learn discriminative distribution features , we incorporate the \textit{distribution loss} which plays a significant role in contributing to faster and better convergence.
We define the \textit{distribution loss} for generation $l$ as follows:
\begin{align}
    \mathcal{L}_l^d=\mathcal{L}_{CE}(\operatorname{Softmax}  (\sum_{j=1}^{N K} {e^d_{l,ij}} \cdot \operatorname{\textit{one-hot}}(y_j)) , {y_i}) \,,
\end{align}
where $e^d_{l,ij}$ stands for the edge feature in the distribution graph at generation $l$. 

The total objective function is a weighted summation of all the
losses mentioned above:
\begin{align}
    \mathcal{L} = \sum_{l=1}^{\hat{l}}(\lambda_p \mathcal{L}_l^p + \lambda_d\mathcal{L}_l^d)\,,
\end{align}
where $\hat{l}$ denotes total generations of DPGN and the weights $\lambda_p$ and $\lambda_d$
of each loss are set to balance their importance. In most of our experiments, $\lambda_p$ and $\lambda_d$ are set to 1.0 and 0.1 respectively.

\section{Experiments}
\subsection{Datasets and Setups}
\subsubsection{Datesets} 
We evaluate DPGN on four standard few-shot learning benchmarks: \textit{mini}ImageNet \cite{vinyals2016matching}, \textit{tiered}ImageNet \cite{ren2018meta}, CUB-200-2011 \cite{wah2011caltech} and  CIFAR-FS \cite{bertinetto2018metalearning}. The \textit{mini}ImageNet and \textit{tiered}ImageNet are the subsets of ImageNet \cite{russakovsky2015imagenet}. CUB-200-2011 is initially designed for fine-grained classification and CIFAR-FS is a subset of CIFAR-100 for few-shot classification. As shown in Table \ref{dat}, we list details for images number, classes number, images resolution and train/val/test splits following the criteria of previous works \cite{vinyals2016matching, ren2018meta, chen19closerfewshot, bertinetto2018metalearning}.

\begin{table}[!h]
\centering
\caption{Details for few-shot learning benchmarks.}
\label{dat}
\setlength{\tabcolsep}{0.8mm}{
\begin{tabular}{ccccc}
\hline
 Dataset& Images & Classes &  Train-val-test & Resolution \\
\hline
\textit{mini}ImageNet & 60000 & 100 & 64/16/20&84x84  \\ 
\hline
\textit{tiered}ImageNet & 779165 & 608 & 351/97/160&84x84 \\ 
\hline
 CUB-200-2011 & 11788 & 200 & 100/50/50&84x84 \\ 
\hline
 CIFAR-FS & 60000 & 100 & 64/16/20 &32x32 \\ 
\hline

\end{tabular}}
\end{table}

\subsubsection{Experiment Setups}
\paragraph {Network Architecture} We use four popular networks for fair comparison, which are ConvNet, ResNet12, ResNet18 and WRN that are used in EGNN \cite{kim2019edge}, MetaOptNet \cite{lee2019meta}, CloserLook \cite{chen19closerfewshot} and LEO \cite{rusu2018meta} respectively.
ConvNet mainly consists of four Conv-BN-ReLU blocks. The last two blocks also contain a dropout layer \cite{srivastava2014dropout}.
ResNet12 and ResNet18 are the same as the one described in \cite{he2016deep}. They mainly have four blocks, which include one residual block for ResNet12 and two residual blocks for ResNet18 respectively.
WRN was firstly proposed in \cite{zagoruyko2016wide}. It mainly has three residual blocks and the depth of the network is set to 28 as in \cite{rusu2018meta}.
The last features of all backbone networks are processed by a global average pooling, then followed by a fully-connected layer with batch normalization \cite{ioffe2015batch} to obtain a 128-dimensions instance embedding. 

\paragraph{Training Schema} We perform data augmentation before training, such as horizontal flip, random crop, and color jitter (brightness, contrast, and saturation), which are mentioned in \cite{gidaris2018dynamic, ye2018learning}. We randomly sample 28 meta-task episodes in each iteration for meta-training. The Adam optimizer is used in all experiments with the initial learning rate of $10^{-3}$. We decay the learning rate by 0.1 per 15000 iterations and set the weight decay to $10^{-5}$.

\paragraph{Evaluation Protocols} We evaluate DPGN in 5way-1shot/5shot settings on standard few-shot learning datasets, \textit{mini}ImageNet, \textit{tiered}ImageNet, CUB-200-2011 and CIFAR-FS. We follow the evaluation process of previous approaches \cite{kim2019edge, rusu2018meta, ye2018learning}. We randomly sample 10,000 tasks then report the mean accuracy (in \%) as well as the 95\% confidence interval.

\subsection{Experiment Results}
\paragraph{Main Results} We compare the performance of DPGN with several state-of-the-art models including graph and non-graph methods. For fair comparisons, we employ DPGN on \textit{mini}ImageNet, \textit{tiered}ImageNet, CIFAR-FS and CUB-200-2011 datasets, which is compared with other methods in the same backbones.
As shown in Table \ref{mini}, \ref{tiered} and \ref{cub}, the proposed DPGN is superior to other existing methods and achieves the state-of-the-art performance, especially compared with the graph-based methods.

\begin{table}[!h]
\centering 
\caption{Few-shot classification accuracies on \textit{\textit{mini}ImageNet}. $^{\dagger}$ denotes thatit is implemented by public code. \cite{garcia2017few, liu2018learning, kim2019edge} and DPGN are tested in transduction.}
\label{mini}
\setlength{\tabcolsep}{0.8mm}{
\begin{tabular}{cccc}

\hline
Method & Backbone & {5way-1shot} & {5way-5shot} \\ \hline
MatchingNet \cite{vinyals2016matching}             & ConvNet                    & 43.56$\pm$0.84                     & 55.31$\pm$ 0.73                   \\
ProtoNet \cite{snell2017prototypical}                & ConvNet                    & 49.42$\pm$0.78                     & 68.20$\pm$0.66               \\
RelationNet \cite{sung2018learning}            & ConvNet                    & 50.44$\pm$0.82                     & 65.32$\pm$0.70                \\
R2D2 \cite{bertinetto2018metalearning}                  & ConvNet                    & 51.20$\pm$0.60                       & 68.20$\pm$0.60                   \\
MAML \cite{finn2017model}                   & ConvNet                    & 48.70$\pm$1.84                     & 55.31$\pm$0.73                \\
Dynamic \cite{gidaris2018dynamic}       & ConvNet                    & 56.20$\pm$0.86                     & 71.94$\pm$0.57                    \\
GNN \cite{garcia2017few}                   & ConvNet                    & 50.33$\pm$0.36                          & 66.41$\pm$0.63                           \\
TPN \cite{liu2018learning}                & ConvNet                    & 55.51$\pm$0.86                           & 69.86$\pm$0.65         \\
Global \cite{luo2019few} &ConvNet&53.21$\pm$0.40&72.34$\pm$0.32\\
Edge-label \cite{kim2019edge}           & ConvNet                    & 59.63$\pm$0.52$^{\dagger}$                           & 76.34$\pm$0.48                        \\
\textbf{DPGN}           & \textbf{ConvNet}                    & \textbf{66.01$\pm$0.36}             & \textbf{82.83$\pm$0.41}       \\\hline
LEO \cite{rusu2018meta}                 & WRN                 & 61.76$\pm$0.08                     & 77.59$\pm$0.12              \\
wDAE \cite{gidaris2019generating} & WRN & 61.07$\pm$0.15& 76.75$\pm$0.11\\
\textbf{DPGN}           & \textbf{WRN}                 & \textbf{67.24$\pm$0.51}             & \textbf{83.72$\pm$0.44}           \\\hline
CloserLook \cite{chen19closerfewshot} &ResNet18&51.75$\pm$0.80&74.27$\pm$0.63\\
CTM \cite{li2019finding}                    & ResNet18                  & 62.05$\pm$0.55                     & 78.63$\pm$0.06             \\
\textbf{DPGN}           & \textbf{ResNet18}                  & \textbf{66.63$\pm$0.51}            & \textbf{84.07$\pm$0.42}           \\\hline
MetaGAN \cite{zhang2018metagan}              & ResNet12                  & 52.71$\pm$0.64                     & 68.63$\pm$0.67                   \\
SNAIL \cite{mishra2017simple}               & ResNet12                  & 55.71$\pm$0.99                     & 68.88$\pm$0.92                   \\
TADAM \cite{oreshkin2018tadam}                   & ResNet12                  & 58.50$\pm$0.30                     & 76.70$\pm$0.30                    \\
Shot-Free \cite{ravichandran2019few} &ResNet12&59.04$\pm$0.43&77.64$\pm0.39$\\
Meta-Transfer \cite{sun2019mtl}   & ResNet12                  & 61.20$\pm$1.80                       & 75.53$\pm$0.80                     \\
FEAT \cite{ye2018learning}            & ResNet12                     & 62.96$\pm$0.02                     & 78.49$\pm$0.02                    \\
TapNet \cite{yoon2019tapnet}              & ResNet12                  & 61.65$\pm$0.15                     & 76.36$\pm$0.10               \\
Dense \cite{lifchitz2019dense} &ResNet12 &62.53$\pm$0.19 &78.95$\pm$0.13\\
MetaOptNet \cite{lee2019meta}            & ResNet12                  & 62.64$\pm$0.61                     & 78.63$\pm$0.46               \\
\textbf{DPGN}           & \textbf{ResNet12}                  & \textbf{67.77$\pm$0.32}             & \textbf{84.60$\pm$0.43}      \\\hline
\end{tabular}}
\end{table}

\begin{table}[h!]
\caption{Few-shot classification accuracies on  \textit{\textit{tiered}ImageNet}. $^{\dagger}$ denotes that it is implemented by public code. * denotes that it is reported from \cite{lee2019meta}. \cite{liu2018learning, kim2019edge} and DPGN are tested in transduction.}
\label{tiered}
\setlength{\tabcolsep}{0.8mm}{
\begin{tabular}{cccc}

\hline
Method & backbone & 5way-1shot & 5way-5shot \\ \hline
MAML* \cite{finn2017model} & ConvNet & 51.67$\pm$1.81 & 70.30$\pm$1.75 \\
ProtoNet* \cite{snell2017prototypical} & ConvNet & 53.34$\pm$0.89 & 72.69$\pm$0.74 \\
RelationNet* \cite{sung2018learning} & ConvNet & 54.48$\pm$0.93 & 71.32$\pm$0.78 \\
TPN \cite{liu2018learning}  & ConvNet & 59.91$\pm$0.94 & 73.30$\pm$0.75 \\
Edge-label \cite{kim2019edge}  & ConvNet & 63.52$\pm$0.52$^{\dagger}$ & 80.24$\pm$0.49 \\
\textbf{DPGN} & \textbf{ConvNet} & \textbf{69.43$\pm$0.49} & \textbf{85.92$\pm$0.42} \\\hline
CTM \cite{li2019finding}  & ResNet18 & 64.78$\pm$0.11 & 81.05$\pm$0.52 \\
\textbf{DPGN} & \textbf{ResNet18} & \textbf{70.46$\pm$0.52} & \textbf{86.44$\pm$0.41} \\\hline
TapNet \cite{yoon2019tapnet} & ResNet12 & 63.08$\pm$0.15 & 80.26$\pm$0.12 \\
Meta-Transfer \cite{sun2019mtl}   & ResNet12  & 65.62$\pm$1.80$^{\dagger}$ & 80.61$\pm$0.90$^{\dagger}$   \\
MetaOptNet \cite{lee2019meta}  & ResNet12 & 65.81$\pm$0.74 & 81.75$\pm$0.53 \\
Shot-Free \cite{ravichandran2019few} &ResNet12&66.87$\pm$0.43&82.64$\pm$0.39\\
\textbf{DPGN} & \textbf{ResNet12} & \textbf{72.45$\pm$0.51} & \textbf{87.24$\pm$0.39} \\\hline
\end{tabular}}
\end{table}

\begin{table}[h!]
\centering
\caption{Few-shot classification accuracies on CUB-200-2011 and CIFAR-FS. * denotes that it is reported from \cite{lee2019meta} or \cite{chen19closerfewshot}. DPGN are tested in transduction.}
\label{cub}
\setlength{\tabcolsep}{1.1mm}{
\begin{tabular}{cccc}

\hline
\multirow{2}{*}{Method} & \multirow{2}{*}{backbone}& \multicolumn{2}{c}{\textbf{CUB-200-2011}} \\ \cline{3-4} 
 && 5way-1shot & 5way-5shot \\ \hline
ProtoNet* \cite{snell2017prototypical}& ConvNet & 51.31$\pm$0.91 & 70.77$\pm$0.69 \\
MAML* \cite{finn2017model}& ConvNet& 55.92$\pm$0.95 & 72.09$\pm$0.76 \\
MatchingNet* \cite{vinyals2016matching}&ConvNet& 61.16$\pm$0.89 & 72.86$\pm$0.70 \\
RelationNet* \cite{sung2018learning} &ConvNet& 62.45$\pm$0.98 & 76.11$\pm$0.69 \\
CloserLook \cite{chen19closerfewshot}&ConvNet& 60.53$\pm$0.83& 79.34$\pm$0.61\\
DN4 \cite{li2019revisiting} &ConvNet& 53.15$\pm$0.84 & 81.90$\pm$0.60 \\
\textbf{DPGN} &\textbf{ConvNet}& \textbf{76.05$\pm$0.51} & \textbf{89.08$\pm$0.38} \\\hline
FEAT \cite{ye2018learning}   &ResNet12& 68.87$\pm$0.22 & 82.90$\pm$0.15 \\
\textbf{DPGN} &\textbf{ResNet12}& \textbf{75.71$\pm$0.47} & \textbf{91.48$\pm$0.33} \\\hline
\multirow{2}{*}{Method} &\multirow{2}{*}{backbone}& \multicolumn{2}{c}{\textbf{CIFAR-FS}} \\ \cline{3-4} 
 && 5way-1shot & 5way-5shot \\ \hline
ProtoNet* \cite{snell2017prototypical}&ConvNet & 55.5$\pm$0.7 & 72.0$\pm$0.6 \\
MAML* \cite{finn2017model} &ConvNet &58.9$\pm$1.9 & 71.5$\pm$1.0 \\
RelationNet* \cite{sung2018learning}& ConvNet& 55.0$\pm$1.0 & 69.3$\pm$0.8 \\ 
R2D2 \cite{bertinetto2018metalearning} &ConvNet& 65.3$\pm$0.2 & 79.4$\pm$0.1\\
\textbf{DPGN} &\textbf{ConvNet}& \textbf{76.4$\pm$0.5}  & \textbf{88.4$\pm$0.4} \\\hline
Shot-Free \cite{ravichandran2019few} &ResNet12&69.2$\pm$0.4&84.7$\pm$0.4\\
MetaOptNet \cite{lee2019meta}& ResNet12& 72.0$\pm$0.7&84.2$\pm$0.5 \\
\textbf{DPGN} &\textbf{ResNet12}& \textbf{77.9$\pm$0.5} & \textbf{90.2$\pm$0.4} \\\hline 
\end{tabular}}
\vspace{-1.0em}
\end{table}

\begin{figure}[h!]

\centering
    \includegraphics[width=8cm]{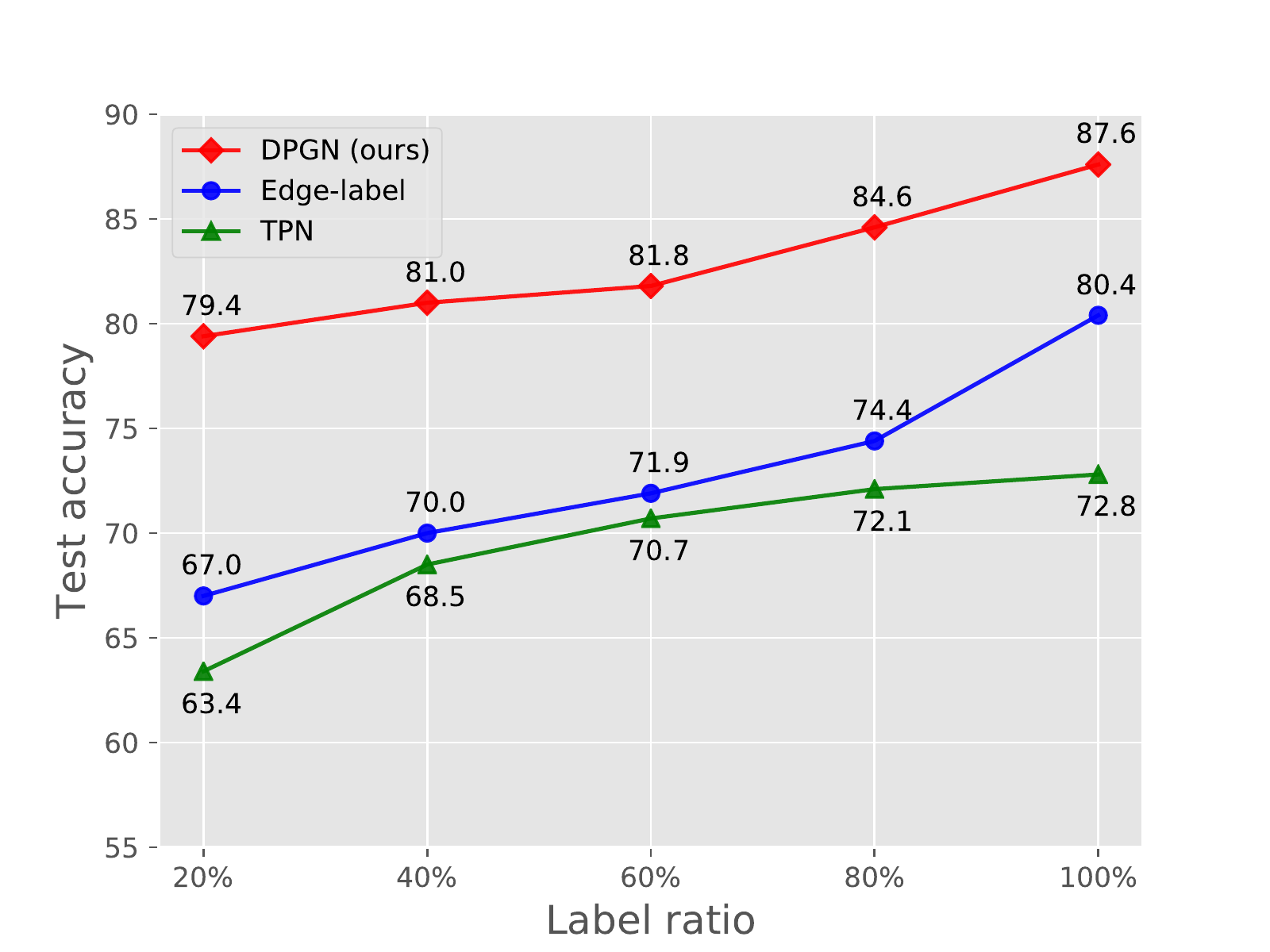}
    
\caption{Semi-supervised few-shot learning accuracy in 5way-10shot on \textit{mini}ImageNet. DPGN surpass TPN and EGNN by a large margin consistently.}
\label{fig:f2}
\end{figure}
\vspace{-1.0em}

\paragraph{Semi-supervised Few-shot Learning}We employ DPGN on semi-supervised few-shot learning. Following \cite{liu2018learning, kim2019edge}, we use the same criteria to split \textit{mini}ImageNet dataset into labeled and unlabeled parts with different ratios. For a 20\% labeled semi-supervised scenario, we split the support samples with a ratio of 0.2/0.8 for labeled and unlabeled data in each class. In semi-supervised few-shot learning, DPGN uses
unlabeled support samples to explicitly construct similarity distributions over all other samples and the distributions work as a connection between queries and labeled support samples, which could propagate label information from labeled samples to queries sufficiently.   
\begin{table}[h!]
\centering
\caption{Trasductive/non-transductive experiments on  \textit{mini}ImageNet. ``BN'' means information is shared among test examples using batch normalization. $^{\dagger}$ denotes that it is implemented by public code released by authors.}
\label{trans}
\setlength{\tabcolsep}{1.1mm}{
\begin{tabular}{ccc}
\hline
Method & Transduction & 5way-5shot  \\ \hline
Reptile \cite{nichol2018first}      &No & 62.74\\
GNN \cite{garcia2017few}         & No & 66.41 \\
Edge-label \cite{kim2019edge}           & No & 66.85$^{\dagger}$\\
\textbf{DPGN}           & \textbf{No} &\textbf{72.83}\\\hline

MAML \cite{finn2017model}           & BN & 63.11\\
Reptile \cite{nichol2018first}     &BN& 65.99\\
RelationNet \cite{sung2018learning}    &BN& 67.07\\\hline

MAML \cite{finn2017model} &Yes & 66.19 \\
TPN \cite{liu2018learning}            &Yes & 69.86\\
Edge-label \cite{kim2019edge} &Yes & 76.37\\
\textbf{DPGN} &\textbf{Yes} & \textbf{84.62}\\\hline
\end{tabular}}
\end{table}

In Figure \ref{fig:f2}, DPGN shows the superiority to exsisting semi-supervised few-shot methods and the result demonstrates the effectiveness to exploit the relations between labeled and unlabeled data when the label ratio decreases.
Notably, DPGN surpasses TPN \cite{liu2018learning} and EGNN \cite{kim2019edge} by 11\% $\sim$ 16\% and 7\% $\sim$ 13\% respectively in few-shot average classification accuracy on \textit{mini}ImageNet.

\begin{figure}[h!]
\centering
    \includegraphics[width=8cm]{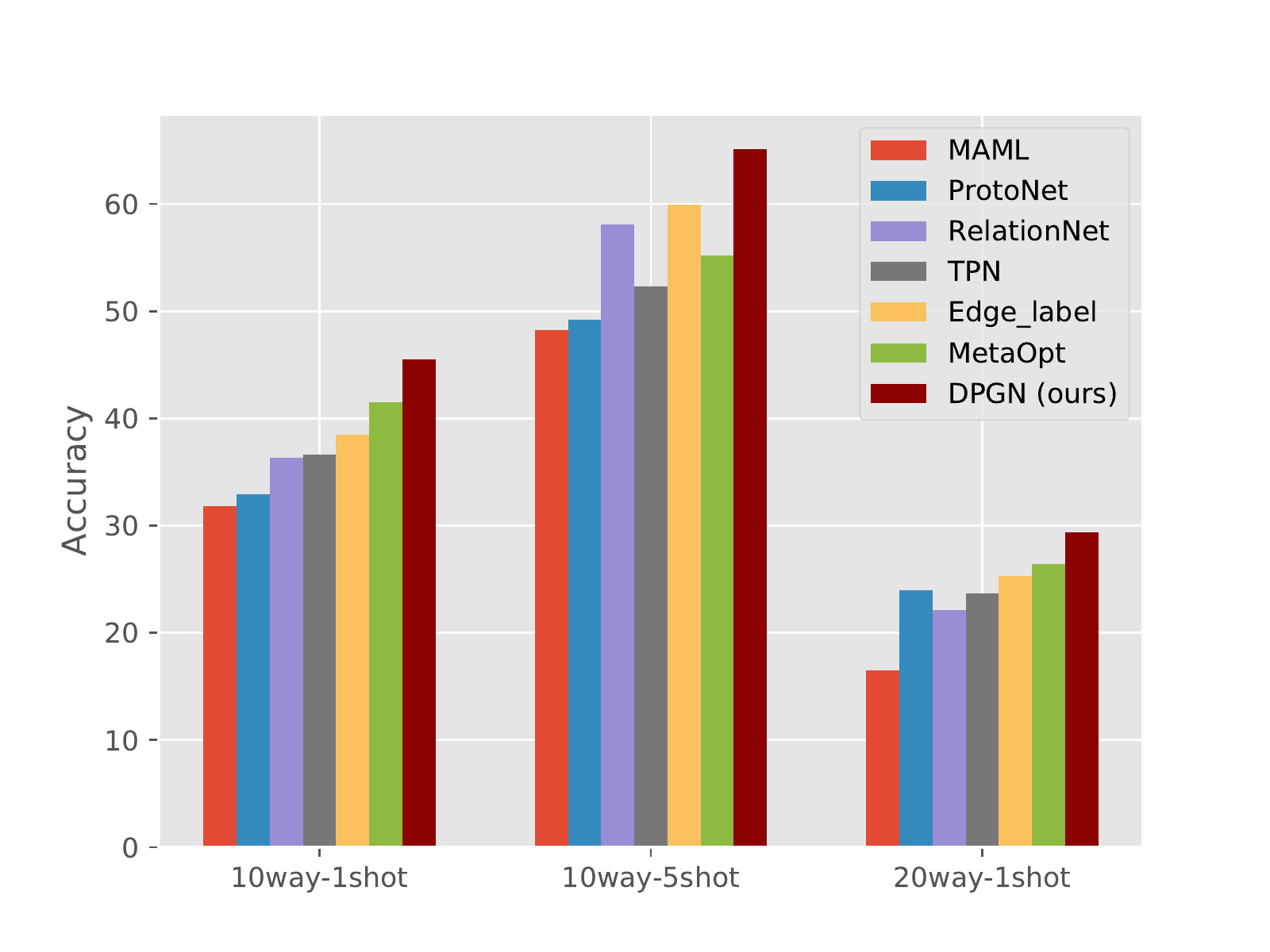}
    
\caption{High-way few-shot classification accuracies on \textit{mini}ImageNet.}
\label{fig:f9}
\end{figure}

\paragraph{Transductive Propagation} To validate the effectiveness of the transductive setting in our framework, we conduct the transductive and non-transductive experiments on \textit{mini}ImageNet dataset in 5way-5shot setting. Table \ref{trans} shows that the accuracy of DPGN increases by a large margin in the transductive setting (comparing with non-transductive setting). Compared to TPN and EGNN which consider instance-level features only, DPGN utilizes distribution similarities between query samples and adopts dual graph architecture to propagate label information in a sufficient way.

\paragraph{High-way classification} Furthermore, the performance of DPGN in high-way few-shot scenarios is evaluated on \textit{mini}ImageNet dataset and its results are shown in Figure \ref{fig:f9}. The observed results show that DPGN not only exceeds the powerful graph-based methods \cite{liu2018learning,kim2019edge} but also surpasses the state-of-the-art non-graph methods significantly. As the number of ways increasing in few-shot tasks, it can broaden the horizons of distribution utilization and make it possible for DPGN to collect more abundant distribution-level information for queries.




\subsection{Ablation Studies}

\begin{figure}[h!]
\centering
    \includegraphics[width=8cm]{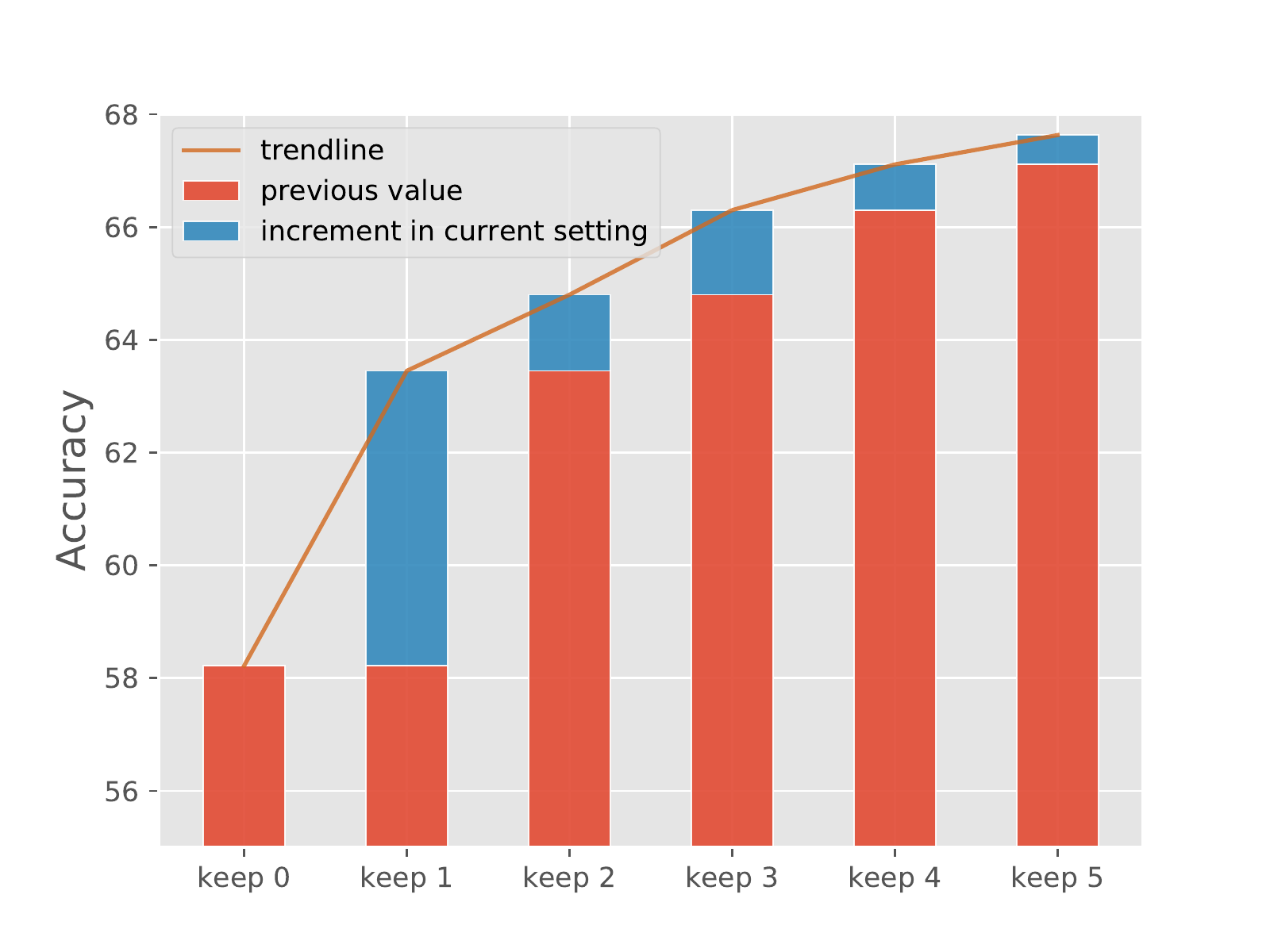}


\caption{Effectiveness of $G^d_l$ through keeping n dimensions in 5way-1shot on \textit{mini}ImageNet.}
\label{fig:info}
\vspace{-1.0em}
\end{figure}

\begin{figure}[h!]

\centering

    \includegraphics[width=8cm]{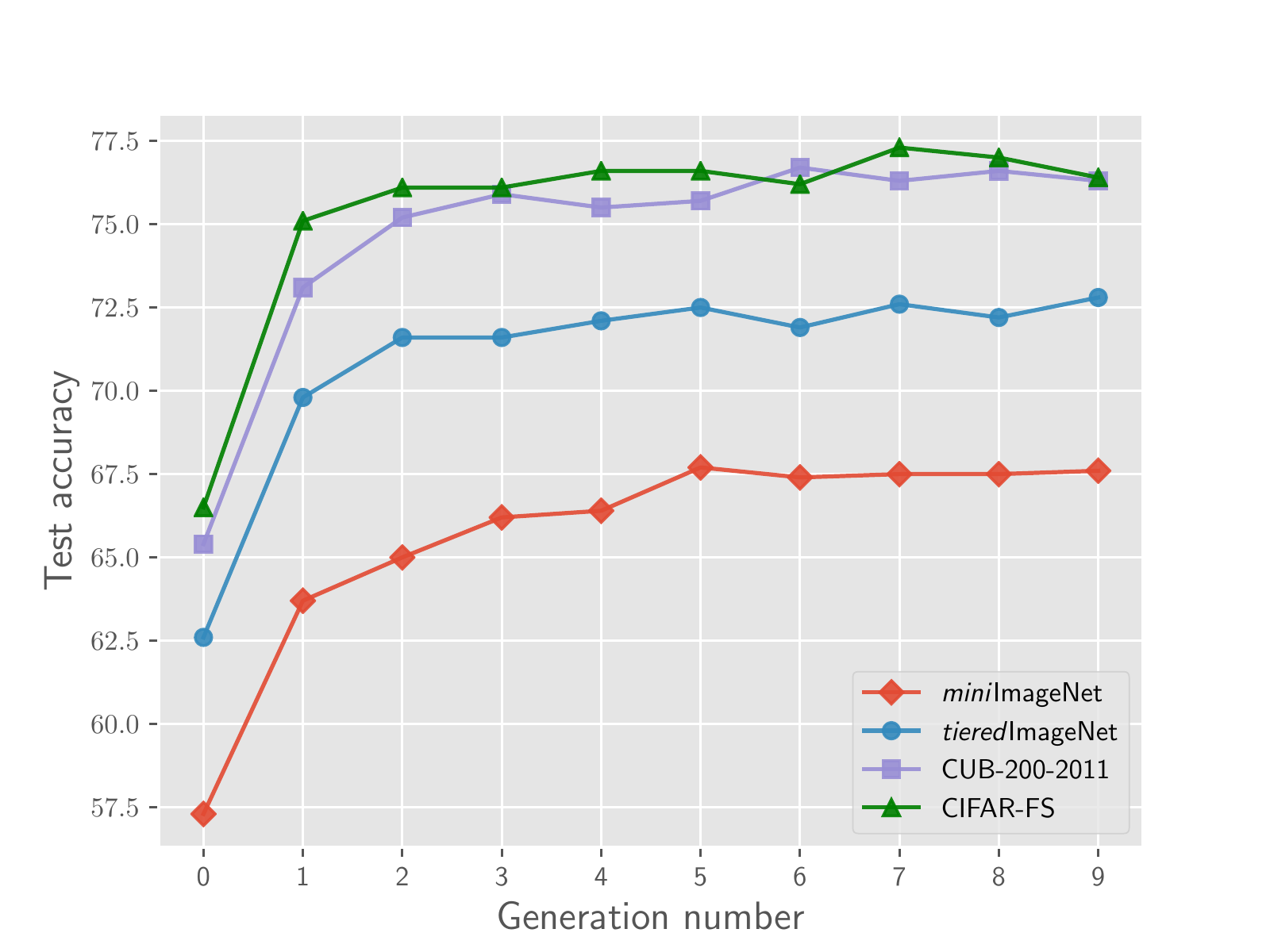}

\caption{Generation number in DPGN on \textit{mini}ImageNet, \textit{tiered}ImageNet, CUB-200-2011 and  CIFAR-FS.}
\label{fig:circle}
\end{figure}

\paragraph {Impact of Distribution Graph} The distribution graph $G^d_l$ works as an important component of DPGN by propagating distribution information, so it is necessary to investigate the effectiveness of $G^d_l$ quantitatively. We design the experiment by limiting the distribution similarities which flow to $G^p_l$ for performing aggregation in each generation during the inference process. Specifically, we mask out the edge features $E^d_l$ through keeping a different number of feature dimensions
and set the value of rest dimensions to zero, since zero gives no contribution. Figure \ref{fig:info} shows the result for our experiment in 5way-1shot on \textit{mini}ImageNet. It is obvious that test accuracy and the number of feature dimensions kept in $E^d_l$ have positive correlations and accuracy increment (area in blue) decreases with more feature dimensions. Keeping dimensions from $0$ to $5$, DPGN boosts the performance nearly by 10\% in absolute value and the result shows that the distribution graph has a great impact on our framework.

\begin{figure}[h]

\centering

    \includegraphics[width=8cm]{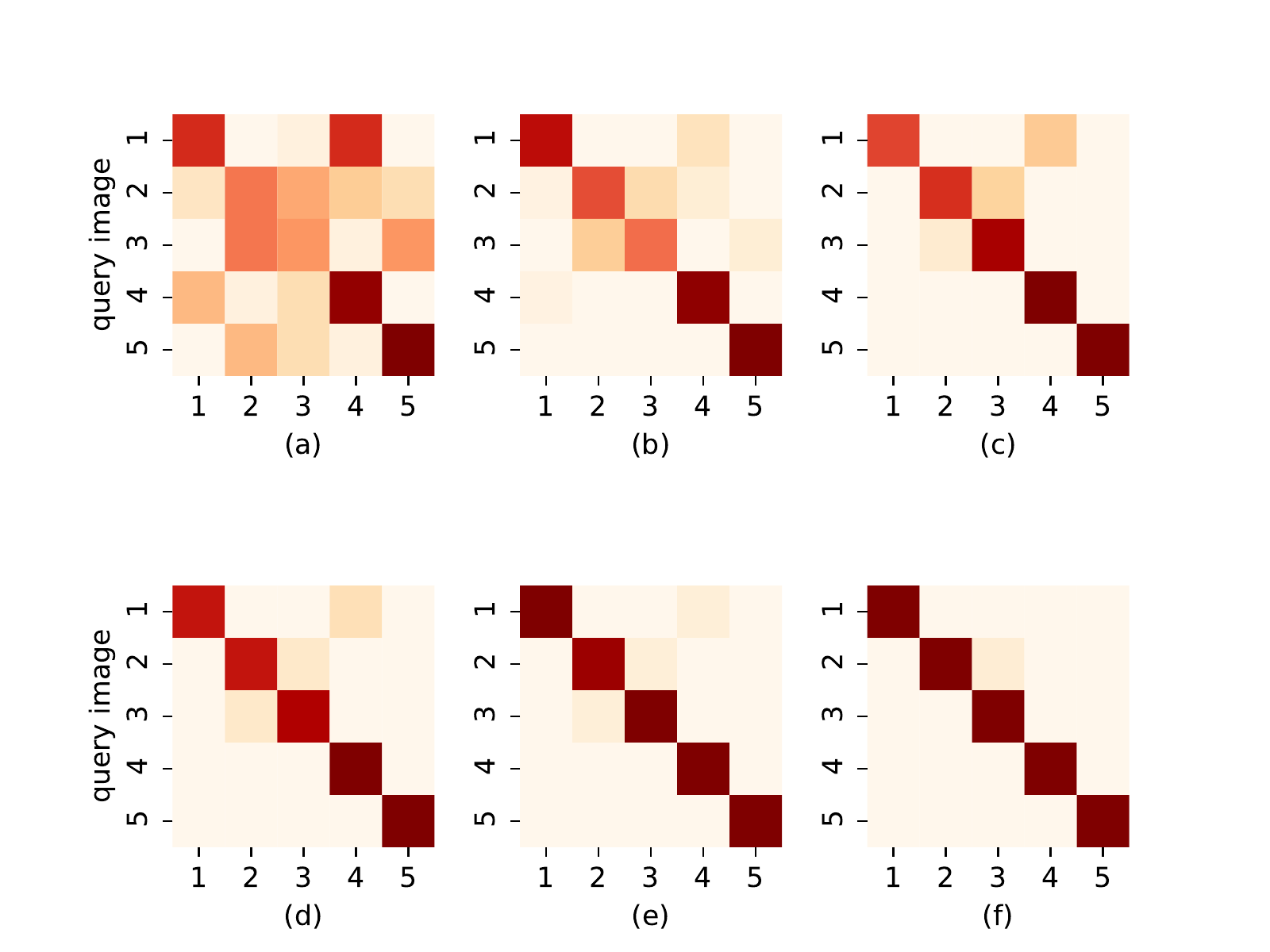}

\caption{The visualization of edge prediction in each generation of DPGN. (a) to (f) denotes generation  1 to 6. The dark denotes higher score and the shallow denotes lower confidence. The left axis stands for the index of 5 query images and the bottom axis stands for 5 support class.}
\label{fig:heatmap}
\vspace{-1.5em}  
\end{figure}



\vspace{-0.5em}

\paragraph {Generation Numbers} DPGN has a cyclic architecture that includes point graph and distribution graph, each graph has node-update and edge-update modules respectively. The total number of generations is an important ingredient for DPGN,
so we perform experiments to obtain the trend of test accuracy with different generation numbers in DPGN on \textit{mini}ImageNet, \textit{tiered}ImageNet, CUB-200-2011, and  CIFAR-FS.
In Figure \ref{fig:circle}, with the generation number changing from 0 to 1, the test accuracy has a significant rise. When the generation number changes from 1 to 10, the test accuracy increases by a small margin and the curve becomes to fluctuate in the last several generations. Considering that more generations need more iterations to converge, we choose generation 6 as a trade-off between the test accuracy and convergence time. 
Additionally, to visualize the procedure of cyclic update, we choose a test scenario where the ground truth classes of five query images are [1, 2, 3, 4, 5] and visualize instance-level similarities which is used for predictions of five query samples as shown in Figure \ref{fig:heatmap}. The heatmap shows DPGN refines the instance-level similarity matrix after several generations and makes the right predictions for five query samples in the final generation. Notably, DPGN not only contributes to predicting more accurately but also enlarge the similarity distances between the samples in different classes through making instance features more discriminative, which cleans the prediction heatmap.

\vspace{-0.5em} 

\section{Conclusion}

In this paper, we have presented the Distribution Propagation Graph Network for few-shot learning, a dual complete graph network that combines instance-level and distribution-level relations in an explicit way equipped with label propagation and transduction. The point and distribution losses are used to jointly update the parameters of the DPGN with \textit{episodic training}. Extensive experiments demonstrate that our method outperforms recent state-of-the-art algorithms by 5\%$\sim$12\% in the supervised task and  7\%$\sim$13\% in semi-supervised task on few-shot learning benchmarks. For future work,  we aim to focus on the high-order message propagation through encoding more complicated information which is linked with task-level relations.

\vspace{-0.5em}

\section{Acknowledgement}
This research was supported by National Key R\&D Program of China (No. 2017YFA0700800).

{\small
\bibliographystyle{ieee_fullname}
}

\end{document}